# Soft Set Based Feature Selection Approach for Lung Cancer Images

Jothi. G and Hannah Inbarani. H

**Abstract**— Lung cancer is the deadliest type of cancer for both men and women. Feature selection plays a vital role in cancer classification. This paper investigates the feature selection process in Computed Tomographic (CT) lung cancer images using soft set theory. We propose a new soft set based unsupervised feature selection algorithm. Nineteen features are extracted from the segmented lung images using gray level co-occurence matrix (GLCM) and gray level different matrix (GLDM). In this paper, an efficient Unsupervised Soft Set based Quick Reduct (SSUSQR) algorithm is presented. This method is used to select features from the data set and compared with existing rough set based unsupervised feature selection methods. Then K-Means and Self Organizing Map (SOM) clustering algorithms are used to cluster the data. The performance of the feature selection algorithms is evaluated based on performance of clustering techniques. The results show that the proposed method effectively removes redundant features.

——————————— ◆ ———————————

## 1 INTRODUCTION

THE second most common malignant (cancerous) tumor is lung cancer. Each year, more people die of lung cancer than of breast, colon, and prostate cancers combined. Early time diagnosis of the lung cancer's pathological type could improve patients' treatment effect. Recently, CT-scanning has emerged as one of the essential diagnostic measures of lung cancer, due to its pattern recognition, machine learning and image process technology [1]. Feature Selection is an essential part of knowledge discovery. FS is used to improve the classification accuracy and reduce the computational time of classification algorithms. FS is divided into the supervised and unsupervised categories. When class labels of the data are available we use supervised feature selection, otherwise unsupervised feature selection is appropriate. In many data mining applications, class labels are unknown, thereby indicating the significance of unsupervised feature selection [2]. In terms of feature selection methods, they fall into filter and wrapper categories. In filter model, features are evaluated based on the general characteristics of the data without relying on any mining algorithms. On the contrary, wrapper model requires one mining algorithm and uses its performance to determine the goodness of feature sets [3]. Soft set theory was first proposed by D. Molodtsov in 1999 for dealing with uncertainties [4]. Soft Set Theory has been applied to data analysis and decision support system. Soft set is trouble-free for attribute reduction in Boolean-value information system. The proposed work exploits soft set theory based operations for feature reduction and then it is compared with rough set based unsupervised algorithms. This soft set based method selects the minimal set of attributes when compared with Rough Set. The main purpose of the proposed algorithm is to increase the efficiency of feature selection method.

The rest of the paper is structured as Sections 2 to 8. Section 2 describes the related work of feature reduction and decision making using soft sets. Section 3 describes the research motivation of this work. Section 4 explains the preprocessing steps for lung image. Section 5 describes the preliminaries of soft set theory. Section 6 explains the proposed algorithm with an example. Section 7 analyses the experimental results. Finally, conclusion is given in Section 8.

## 2 RELATED WORK

In [5], a filter-based feature selection method, biomarker identifier (BMI), is adopted to analyze gene expression data that might be used to discriminate between samples with and without lung cancer. Amer et al. [6] investigated computed tomographic (CT) chest images and developed a computer-aided system to discriminate different lung abnormalities. They studied texture based features, fourier-based features and wavelet-based features. Aliferis et al. [7] explores machine learning methods for the development of computer models. These computer models used gene expression data to distinguish between tumor and non-tumor, between metastatic and non-metastatic, and between histological subtypes of lung cancer. A second goal is to identify small sets of gene predictors and study their properties in terms of stability, size, and relation to lung cancer. The idea of attribute reduction and decision making using soft set theory was first proposed by Maji et al. [8]. In [8] decision selection in given attribute is based on the maximal weighted choice value and it is similar to the Pawlak rough reduction. Herawan et al. [9] introduced a new approach for an attribute reduction in multi-valued information system using soft set. AND operation was used for attribute reduction and it was shown that the reducts obtained were equivalent with Pawla's rough reduction. In [10], the soft set theory has been used as feature selec-

————————————————
- **Jothi. G** *is currently pursuing M. Phil in Computer Science at Periyar University, Salem, India. E-mail: jothiys@gmail.com.*
- **H. Hannah Inbarani** *is currently working an Assistant professor, Department of Computer science, Periyar Uinversity, Salem, India. E-mail: hhinba@gmail.com*

tion technique to identify the best features of Traditional Malay musical instrument sounds.

## 3 RESEARCH MORTIVATION

The proliferation of large data sets within many domains poses unprecedented challenges to data mining [11]. Researchers realize that in order to achieve successful data mining, feature selection is an indispensable component [12]. In image processing the feature selection approach takes enormous amount of time to find minimal subset of features. The new researches in this area focus on reducing runtime for the purpose of efficient research. Hence this work proposes an effective and efficient approach to find the reduct set.

## 4 PRELIMINARIES

### 4.1 Soft Set Theory

Throughout this section $U$ refers to an initial universe, E is a set of parameters, P $(U)$ is the power set of $U$ and A ⊆ E [4].
Definition 1: A pair (F, A) is called a soft set over $U$, where F is a mapping given by
$$F: A \rightarrow P(U)$$

### 4.2 Multi-soft Sets

The idea of "multi soft set" is based on a decomposition of a multi-valued information system $S = (U, A, V, f)$, into |A| number of binary-valued information systems $S = (U, A, V_{\{0,1\}}, f)$, where |A| denotes the cardinality of A. Consequently, the |A| binary-valued information systems define multi-soft sets [13].
$$(F, E) = \{(F, a_i): i = 1, 2, \ldots |A|\}$$

### 4.3 AND operation in multi-soft sets

Definition 2: Let $(F, E) = ((F, a_i): i = 1, 2, \ldots |A|)$ be a multi-soft set over U representing a multi-valued information system $S = (U, A, V, f)$. The AND operation between $(F, a_i)$ and $(F, a_j)$ is defined as [9].
$(F, a_i) AND (F, a_j) = (F, a_i \times a_j)$, where
$G(Va_i, Va_j) = F(Va_i) \cap F(Va_j), \forall (Va_i, Va_j)$
$\in a_i \times a_j, for\ 1 \leq i, j \leq |A|$.

### 4.4 Attribute Reduction

Definition 3: Let $(F, E) = ((F, a_i): i = 1, 2, \ldots |A|)$ be a multi-soft set over U representing a multi-valued information system $S = (U, A, V, f)$. A set of attributes B ⊆ A is called a reduct for A if $C_{F(b_2 \times \ldots \times b_{|B|})} = C_{F(a_2 \times \ldots \times a_{|A|})}$ [9].

Definition 4: Assume X ⊆ A is an attribute subset, $x \in A$ is an attribute, the importance of $x$ for X is denoted by $Sig_X(x)$ the definition is, [14] $Sig_X(x) = 1 - |X \cup \{x\}|/|X|$, Where |X| = |IND(X)|. Suppose U/IND(X) = U/X={$X_1$, $X_2$ …$X_n$}, then $|X| = |IND(X)| = \sum_{i=1}^{n} |X_i|^2$. $|X| - |X \cup \{x\}|$ represents the decrement of indiscernibility and also the increment of discernibility as attribute x is added to X. The number of selection methods is originally indiscernible in X, but it is discernible in $X \cup \{x\}$, and the increment of indiscernibility is expressed by
$(|X| - |X \cup \{x\}|)/X = 1 - |X \cup \{x\}|/X$.

## 5 THE PROCESSING MODEL

A typical image processing system generally consists of image acquisition, enhancement, segmentation, feature extraction, feature selection and clustering/classification. Lung image categorization process is depicted in Figure 1.

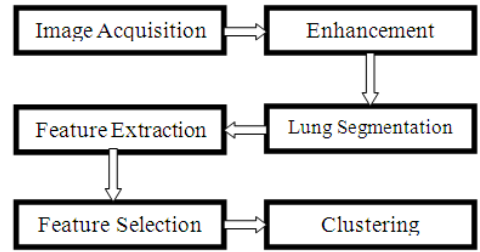

Fig. 1 Lung Image Categorization Process

### 5.1 Image Enhancement

Due to low quality, low ability of distinguishing abnormalities from their surrounding, and artifacts. The quality of CT images could be degraded, sometimes to the point making them diagnostically unusable. So, the first step in preprocessing is the image denoising. Therefore the current step in this work is to perform a comparative study of some image enhancement technique, namely, Average filter, Gaussian filter, Median filter [15].

All mentioned types of filters are applied in the noisy image. The SNR value is obtained using the equation (1) and comparing the values of SNR of each image that resulted from each filter type we get the most suitable filter that could be applied on the raw CT images [6]. The original image and the noisy image are shown in Figure 2 and the resulted images from all types of the previous filters are shown in Figure 3. The calculated SNR of each filtered image for all applied filters are shown in Table 1.

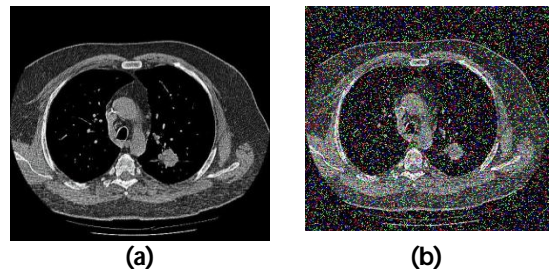

Fig 2. (a) The original image and (b) noisy image.

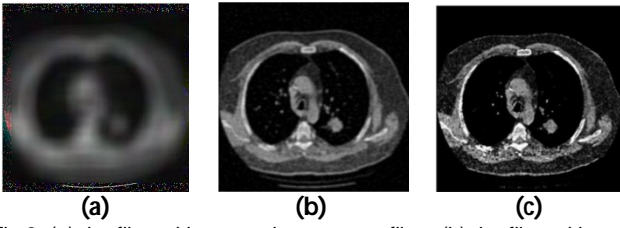

Fig 3. (a) the filtered image using average filter, (b) the filtered image using Gaussian filter, (c) the filtered image using median filter

### 5.1.1 SNR Calculation

$$SNR = 10\log_{10}\left(\sum_{i=1}^{K} S_i^2 / \sum_{i=1}^{K}(\hat{S}_i^2 - S_i^2)\right) \quad \text{------- (1)}$$

Where $\hat{S}$ = Noisy Image, $S$ = Original Image
K = Image Size

TABLE 1.
THE CALCULATED SNR OF THE FILTERED IMAGES USING ALL TYPES OF FILTERS

| Filter Type | SNR Value |
| --- | --- |
| Mean filter | 4.4400 |
| Gaussian filter | 7.5788 |
| Median filter | 11.2069 |

### 5.1.2 Application of Denoising Filters to CT scan Images

The performance of three denoising filter techniques has been compared as above. The results show that the 5×5 Median filter gives the highest quality image compared to the other mentioned filtering techniques. So it is applied to all the raw CT lung images. An example of the filtered CT lung image is shown in Figure 4.

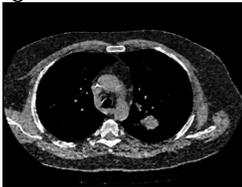

Fig 4. The filtered CT lung image

### 5.2 Lung Segmentation

In this paper, Region Growing Segmentation (RGS) is performed on each image. Figures 5 (a) and (b) demonstrate the performance of before and after RGS algorithm and finally, figure 5 (c) shows the segmented lung image [16].

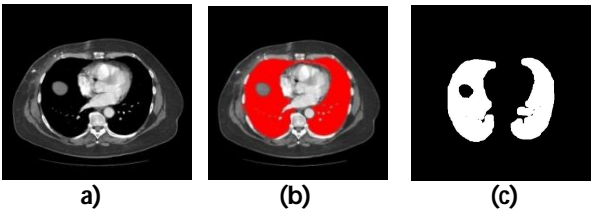

Fig 5. (a) Before RGS (b) After RGS (c) Segmented Image

### 5.3 Feature Extraction

Feature extraction methodologies analyze objects and images to extract the most prominent features that are representative of the various classes of objects. In this research, we used gray level co-occurence matrix GLCM in four possible directions $0^0, 45^0, 90^0$ and $135^0$ and gray level different matrix GLDM to extract texture features from digital CT images [17].

Nineteen texture parameters viz. angular second moment $(f_1)$, contrast$(f_2)$, correlation$(f_3)$, sum of squares: variance$(f_4)$, inverse difference moment$(f_5)$, sum average$(f_6)$, sum variance$(f_7)$, sum entropy$(f_8)$, entropy$(f_9)$, difference variance$(f_{10})$, difference entropy$(f_{11})$, information measures of correlation-I$(f_{12})$, information measures of correlation-II$(f13)$, maximal correlation coefficient$(f_{14})$, cluster shade$(f_{15})$, cluster prominence$(f_{16})$, product moment$(f_{17})$, Inertia$(f_{18})$ and mean$(f_{19})$ are calculated.

## 6 UNSUPERVISED FEATURE SELECTION

In many data mining applications, class labels are unknown, and so considering the significance of Unsupervised Feature Selection (UFS) the proposed work is applied for UFS.

**6.1 USQR Algorithm:** The Unsupervised Quick Reduct (USQR) algorithm attempts to calculate a reduct without exhaustively generating all possible subsets. According to the algorithm, the mean dependency of each attribute subset is calculated and the best feature is chosen. [18]

Algorithm 1: USQR Algorithm - USQR(C)

C, the set of all conditional features;
(1) $R \leftarrow \{\}$
(2) do
(3) T ← $R$
(4) $\forall x \in (C - R)$
(5) $\forall y \in C$
(6) $\gamma_{R \cup \{x\}}(y) = \frac{|POS_{R \cup \{x\}}(y)|}{|U|}$
(7) if $\overline{\gamma_{R \cup \{x\}}(y), \forall y| \in C} > \overline{\gamma_T(y), \forall y| \in C}$
(8) $T \leftarrow R \cup \{x\}$
(9) $R \leftarrow T$
(10) until $\overline{\gamma_{R \cup \{x\}}(y), \forall y| \in C} = \overline{\gamma_T(y), \forall y| \in C}$
(11) return R

**6.1 URR Algorithm:** The Unsupervised Relative Reduct (URR) algorithm starts by considering all the features contained in the dataset. Each feature is then examined iteratively, and the relative dependency measure is calculated. If the relative dependency is equal to one then that feature can be removed. This process continues until all features have been examined [18].

Algorithm 2: URR Algorithm - URR(C)

URR(C)
C, the Conditional attributes
(1) $R \leftarrow C$

(2) $\forall a \in C$
(3)     $if\ (k_{R-\{a\}}(\{a\}) == 1)$
(4)        $R \leftarrow R - \{a\}$
(5) return R

**6.1 SSUSQR Algorithm: - The Proposed Approach:** In the new Soft Set based Unsupervised Quick Reduct algorithm, the dimensionality reduction is achieved by using AND operation in soft set theory. It starts with an empty set and finds the cardinality of the indiscernibility of the universal set. For every conditional feature, the cardinality of indiscernibility is computed. The attribute which has highest cardinality value is taken as the core attribute in the reduct set. If more than one attribute has maximum cardinality value, Sig(x) is found out and that value is taken as the core attribute. In the next step, combination of other attributes with CORE(x) attribute is taken as feature subset and the feature subset with maximum cardinality of indiscernibility is taken. This process continues until the cardinality of indiscernibility of the feature subset is equal to the cardinality of the indiscernibility of the universal set.

Algorithm 2: SSUSQR Algorithm – SSUSQR(C)
U, the Universal Set;
C, the set of all conditional features;
(1) $R \leftarrow \{\}$
(2) Do
(3) $T \leftarrow R$
(4) Find $S_T(U) = |IND(U)|$
(5) $\forall x \in (C-R)$ Find Maximum($|IND(x)|$)
(6) If more than one attribute has same
    Maximum($(|IND(x)|)$)
    Compute significance *Sig* of those attributes
      And Compute $x$ = CORE($x$)
    End if
(7) $T \leftarrow R \cup \{x\}$
(8) $R \leftarrow T$
(9) Until $|R| == S_T(U)$
(10) Return R

6.3. Worked Example: Now, let us consider the given information in table 2. The conditional attributes are {$a_1$, $a_2$, $a_3$, $a_4$}. A multi-value information system as specified in table 3 is constructed.

TABLE 2. INFORMATION TABLE

| U | $a_1$ | $a_2$ | $a_3$ | $a_4$ |
|---|---|---|---|---|
| 1 | Circle | Large | Red | Good |
| 2 | Square | Small | Green | Bad |
| 3 | Triangle | Large | Blue | Bad |
| 4 | Circle | Small | Blue | Good |
| 5 | Circle | Large | Red | Average |
| 6 | Triangle | Medium | Blue | Bad |
| 7 | Triangle | Small | Green | Bad |
| 8 | Square | Small | Green | Average |

TABLE 3. MULTI-VALUE INFORMATION SYSTEM

| U | $a_1$ | | | $a_2$ | | | $a_3$ | | | $a_4$ | | |
|---|---|---|---|---|---|---|---|---|---|---|---|---|
|   | Circle | Square | Triangle | Large | Small | Medium | Red | Green | Blue | Good | Bad | Average |
| 1 | 1 | 0 | 0 | 1 | 0 | 0 | 1 | 0 | 0 | 1 | 0 | 0 |
| 2 | 0 | 1 | 0 | 0 | 1 | 0 | 0 | 1 | 0 | 0 | 1 | 0 |
| 3 | 0 | 0 | 1 | 1 | 0 | 0 | 0 | 0 | 1 | 0 | 1 | 0 |
| 4 | 1 | 0 | 0 | 0 | 1 | 0 | 0 | 0 | 1 | 1 | 0 | 0 |
| 5 | 1 | 0 | 0 | 1 | 0 | 0 | 1 | 0 | 0 | 0 | 0 | 1 |
| 6 | 0 | 0 | 1 | 0 | 0 | 1 | 0 | 0 | 1 | 0 | 1 | 0 |
| 7 | 0 | 0 | 1 | 0 | 1 | 0 | 0 | 1 | 0 | 0 | 1 | 0 |
| 8 | 0 | 1 | 0 | 0 | 1 | 0 | 0 | 1 | 0 | 0 | 0 | 1 |

(F, U) = ((F, $a_1$), (F, $a_2$), (F, $a_3$), (F, $a_4$))

(F, $a_1$) = {{Circle=1, 4, 5}, {Square=2, 8}, {Triangle=3, 6, 7}}
(F, $a_2$) = {{Large=1, 3, 5}, {Small=2, 4, 7, 8}, {Medium=6}}
(F, $a_3$) = {{Red=1, 5}, {Green=2, 7, 8}, {Blue=3, 4, 6}}
(F, $a_4$) = {{Good=1, 4}, {Bad=2, 3, 6, 7}, {Average=5, 8}}

*Step 1:* Begin.
$S_T(U) = \{(F,a_1) AND (F,a_2) AND (F,a_3) AND (F,a_4)\}$
    = F ($a_1 \times a_2 \times a_3 \times a_4$)
$S_T(U) = |\{\{1\}, \{2\}, \{3\}, \{4\}, \{5\}, \{6\}, \{7\}, \{8\}\}| = 8$      ---- (2)

*Step 2:* Calculate the cardinality value.
(F, $a_1$) = |{1, 4, 5}, {2, 8}, {3, 6, 7}| = 3
(F, $a_2$) = 3,    (F, $a_3$) = 3,    (F, $a_4$) = 3.

Find the maximum cardinality, that attribute is the core attribute in the reduct set.

*Step 3:*
In this example, the maximum cardinality value 3 occurs four times. From the Definition 4 we can calculate Significance value. The attribute which has highest Significance value is taken as core attribute CORE (A).

$|A| = S_T(U) = \{\{1\}, \{2\}, \{3\}, \{4\}, \{5\}, \{6\}, \{7\}, \{8\}\}$

$$Sig_{A-\{a1\}}(a_1) = 1 - \frac{1-|A|}{|A-\{a_1\}|}$$

$$= 1 - \frac{1+1+1+1+1+1+1+1}{1+2\times 2+1+1+1+1+1} = 0.2$$

Similarly, we can calculate, $Sig_{A-\{a2\}}(a_2) = 0.2$, $Sig_{A-\{a3\}}(a_3) = 0.2$,    $Sig_{A-\{a4\}}(a_4) = 0.33$.

CORE (A) = {$a_4$}, If $S_T(U) \neq |R|$ then go to next step.

*Step 4:*
Take the combination, {$a_1$, $a_4$}, {$a_2$, $a_4$}, {$a_3$, $a_4$}.
F ($a_1 \times a_4$) = | {1, 4}, {2}, {3, 6, 7}, {5}, {8}| = 5

Similarly, **F ($a_2$ x $a_4$) = 7**,  F ($a_3$ x $a_4$) = 6
Select the attribute which has maximum cardinality.
R = {$a_2$, $a_4$}. If $S_T$ (U) ≠ |R|, then go to next step.

*Step 5:*
Take the combination, {$a_1$, $a_2$, $a_4$}, {$a_2$, $a_3$, $a_4$}.
**F ($a_1$ x $a_2$ x $a_4$) = 8**,     F ($a_2$ x $a_3$ x $a_4$) = 7
Select the attribute which has highest cardinality.
R = {$a_1$, $a_2$, $a_4$}
   $S_T$ (U) == |R|. Then the reduct set is {$a_1$, $a_2$, $a_4$}.

*Step 6:*  End.

## 7   EXPERIMENTAL RESULTS

### 7.1. Data Collection

The Data sets are collected from the National Cancer Institute database and Mitra Scan Centre, Salem [19]. In this experimental analysis, 200 raw CT lung images are taken. The feature extracted table used in this reserach is in the form of continuous value with noncategorical features (attributes). In order to employ the soft set approach proposed by [9, 13], it is essential to transform the dataset into categorical ones. For that, the equal width binning discretization technique in [20] is used.

### 7.2. Feature Reduction

A comparison of the USQR, URR and SSUSQR methods is made based on the subset and clustering performance. The data set name is described according to the textural description matrix and orientation (degree) is used to extract features from the CT lung image. There are 19 different features are extracted and used in our experiment. The selected features are listed in table 4.

Figure 6 demonstrates the efficiency of the feature reduction for our proposed algorithm. It selects the minimal set of features.

TABLE 4.
FEATURES SELECTED USING UNSUPERVISED FEATURE SELECTION METHODS

| Data Set | No. of Features Extracted | No. of Features Selected | | |
|---|---|---|---|---|
| | | USQR | URR | SSUSQR |
| GLCM_0 | 19 | 11 | 10 | 10 |
| GLCM_45 | 19 | 10 | 10 | 9 |
| GLCM_90 | 19 | 10 | 11 | 9 |
| GLCM_135 | 19 | 11 | 10 | 9 |
| GLDM | 19 | 10 | 11 | 9 |

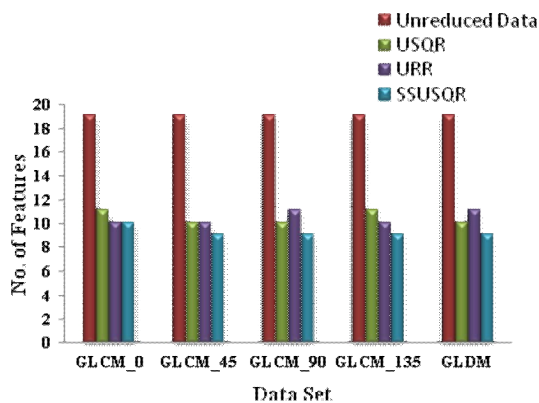

Fig. 6 Reduced set of features for Lung cancer image data set.

### 7.3. Performance Analysis of FS algorithms

The performance of unsupervised feature selection algorithms USQR, URR and SSUSQR are compared before and after feature selection using clustering and cluster validity measures. The feature set is clustered by the k-means and SOM algorithm.

The performance of cluster validity measures Dunn's Index and Silhouette Index are presented in tables 5 and 6. The clustering was initially performed on the un-reduced features followed by reduct features, which were obtained by dimensionality reduction techniques.

TABLE 5.
PERFORMANCE OF K-MEANS CLUSTERING BASED ON CLUSTER VALIDITY MEASURES

| Data Set | Dunn Index | | | | Silhouette Index | | | |
|---|---|---|---|---|---|---|---|---|
| | Un reduced Set | USQR | URR | SSUSQR | Un reduced Set | USQR | URR | SSUSQR |
| GLCM_0 | 1.6039 | 1.5310 | 1.5614 | 1.6575 | 0.3578 | 0.3412 | 0.3492 | 0.4189 |
| GLCM_45 | 1.8796 | 1.6350 | 1.9243 | 2.0308 | 0.4251 | 0.3652 | 0.4348 | 0.4702 |
| GLCM_90 | 1.8146 | 1.6541 | 1.6921 | 2.0311 | 0.4028 | 0.3663 | 0.3765 | 0.4678 |
| GLCM_135 | 1.6466 | 1.5392 | 1.5698 | 1.6488 | 0.3694 | 0.3448 | 0.3828 | 0.3530 |
| GLDM | 2.5475 | 2.6433 | 2.6433 | 2.8801 | 0.6330 | 0.6526 | 0.6526 | 0.7103 |

.

TABLE 6.
PERFORMANCE OF SOM CLUSTERING BASED ON VALIDITY MEASURES

| Data Set | Dunn Index | | | | Silhouette Index | | | |
|---|---|---|---|---|---|---|---|---|
| | Un reduced Set | USQR | URR | SSUSQR | Un reduced Set | USQR | URR | SSUSQR |
| GLCM_0 | 1.4587 | 1.5310 | 1.5614 | 1.6966 | 0.3415 | 0.3413 | 0.3493 | 0.4293 |
| GLCM_45 | 1.6350 | 1.8796 | 1.9243 | 1.9980 | 0.3652 | 0.4251 | 0.4348 | 0.4499 |
| GLCM_90 | 1.8703 | 1.6541 | 1.6921 | 1.3478 | 0.4226 | 0.3663 | 0.3765 | 0.4255 |
| GLCM_135 | 1.6264 | 1.5351 | 1.5672 | 1.7411 | 0.3671 | 0.3411 | 0.3494 | 0.4210 |
| GLDM | 2.5619 | 2.5475 | 2.6433 | 2.8801 | 0.6434 | 0.6330 | 0.6526 | 0.7103 |

Figures 7 (a) and (b) illustrate the performance of Dunn index and Silhouette Index using K-means for the data sets taken based on the features selected using proposed approach. These figures demonstrate the effectiveness of SSUSQR over USQR and URR since it shows higher index value.

demonstrate the effectiveness of SSUSQR over USQR and URR since it shows higher index value.

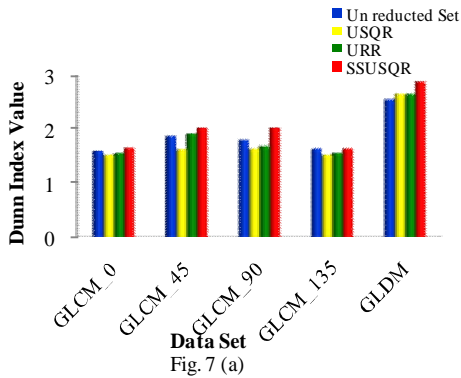

Fig. 7 (a)

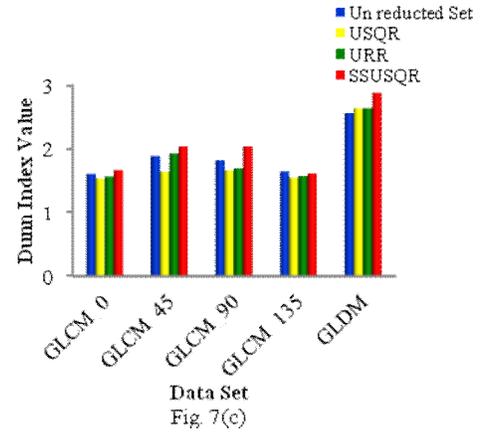

Fig. 7(c)

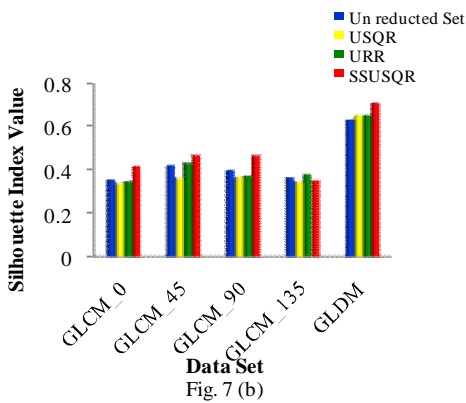

Fig. 7 (b)

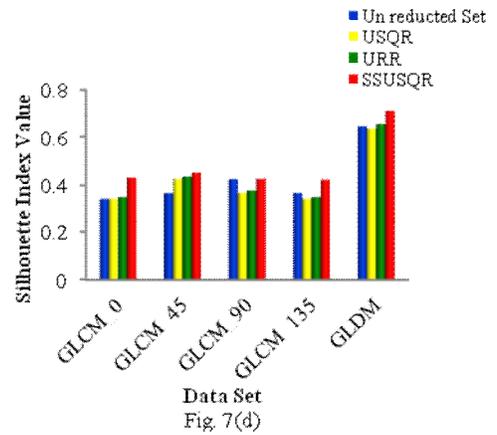

Fig. 7(d)

Figures 7 (c) and (d) show the performance of Dunn index and Silhouette Index using SOM for the data sets taken based on the features selected using proposed approach. These figures

## 8 CONCLUSION

In this paper, USQR, URR and SSUSQR algorithms are analyzed using raw CT lung images. The proposed SSUSQR algorithm using soft set theory effectively removes redundant fea-tures. The selected features are clustered using k-means and SOM clustering algorithms. The Dunn Index and Silhouette index were used for measuring the quality of the clusters obtained. The proposed method provides the best result compared with rough set based unsupervised feature selection. In future, it can be applied to other medical images also.